\pdfoutput=1 

\documentclass[]{interact}

\usepackage{multirow}
\usepackage{natbib}
\bibpunct[, ]{(}{)}{;}{a}{}{,}

\usepackage[bookmarksnumbered,hidelinks]{hyperref}

\theoremstyle{plain}

\theoremstyle{definition}

\theoremstyle{remark}

\begin{document}

\articletype{Research Article}

\title{DWFF-Net: A Multi-Scale Farmland System Habitat Identification Method with Adaptive Dynamic Weight Feature Fusion}

\author{
\name{Kesong Zheng\textsuperscript{a}\thanks{CO-FIRST AUTHOR Kesong Zheng, Zhi Song, Peizhou Li. These authors contributed equally to this work.} , Zhi Song\textsuperscript{b} , Peizhou Li\textsuperscript{c} , Shuyi Yao\textsuperscript{d} , Tong Li\textsuperscript{c} , Yonglin Shen\textsuperscript{a} , and Zhenxing Bian\textsuperscript{d}\thanks{CONTACT Zhenxing Bian. Author. Email: zhx-bian@syau.edu.cn}}
\affil{\textsuperscript{a}National Engineering Research Center of Geographic Information System, China University of Geosciences, Wuhan, China; \textsuperscript{b}College of Science, Shenyang Agricultural University, Liaoning, China;\textsuperscript{c}College of Engineering, Shenyang Agricultural University, Liaoning, China;\textsuperscript{d}College of Land and Environment, Shenyang Agricultural University, Liaoning, China}
}

\maketitle

\begin{abstract}
To address insufficient accuracy in multi-scale segmentation for agricultural habitat recognition, this study proposes a Dynamic Weighted Feature Fusion Network (DWFF-Net). Its encoder uses frozen DINOv3 to extract basic features and introduces a data-level adaptive dynamic weighting strategy based on relationships between image categories and feature maps. The decoder employs a dynamic weight calculation network for deep fusion of multi-level features and a hybrid loss for optimization. Statistical analysis shows that weight entropy tends to decrease as habitat category count increases, indicating adaptive adjustment of fusion strategy according to scene complexity. Experiments on a previously constructed agricultural habitat dataset validate DWFF-Net. Ablations yield mIoU 0.6979 and mF1 0.8049, exceeding the Static Weighted Feature Fusion Network by 1.82\% and 1.54\%, respectively, confirming that dynamic weighting improves multi-level feature utilization. Compared with U-Net, DeepLabv3+, SegFormer, and DPT, DWFF-Net improves mIoU by 16.32\%, 6.49\%, 4.17\%, and 3.18\%, respectively. For tiny features like scattered trees, IoU reaches 0.2707, outperforming those models by 99.85\%, 11.45\%, 22.24\%, and 19.32\%, verifying effectiveness in tiny habitat segmentation. This framework enables low-cost, high-precision habitat mapping and supports refined monitoring in agricultural landscapes.
\end{abstract}


\section{Introduction}

Agricultural landscapes play an irreplaceable role in ensuring food security and maintaining ecological stability \citep{delabreActionsSustainableFood2021,desimonePatternsBiodiversityHabitat2017}. However, the ongoing intensification of agriculture has exacerbated landscape homogenization, leading to a series of global ecological issues such as the degradation of cropland ecological functions and the sharp decline in farmland biodiversity \citep{lindborgFunctionSmallHabitat2014}. Within this context, non-productive spaces---land types not covered by conventional farming activities but that support critical ecological functions, including forest belts, grasslands, ponds, field margins, ditches, roadside edges, and abandoned lands---have emerged as key spatial carriers for enhancing biodiversity and ecosystem services in cropland systems \citep{jingpingConnotationCharacteristicsRecognition2022,tougeronMultiscaleApproachBiodiversity2022}. These areas, collectively referred to as non-cropped habitats, hold significant ecological value; yet their fine-scale identification and dynamic monitoring rely fundamentally on the acquisition and intelligent interpretation of remotely sensed data. Very high-resolution (VHR) remote sensing imagery, in particular, has created unprecedented technological opportunities for detailed recognition and boundary delineation of these small-scale habitats within agricultural landscapes \citep{zhouMachineLearningParadigms2021,zhangFastAccurateLandCover2021,wangParcelScaleCropPlanting2026}, making advanced remote sensing image analysis a critical enabler for quantitative cropland ecological management.

Non-cultivated habitats and cultivated habitats together constitute the cropland system \citep{tianHealthAssessmentDriving2025}. As key spatial carriers for maintaining cropland biodiversity, non-cultivated habitats exhibit rich typological differentiation due to differences in spatial location, geometric shape, vegetation type, and community characteristics \citep{wangDirectIndirectEffects2024a}. Therefore, accurately identifying habitats within the cropland system---particularly distinguishing smaller-scale categories within non-cultivated habitats---is a crucial prerequisite for quantitative research on cropland ecological management \citep{lagerquistManagementinducedMicrohabitatsCrop2025}. From the perspective of remote sensing image analysis, this task translates into a fine-grained semantic segmentation problem that demands precise boundary extraction and pixel-level classification of micro-habitats from VHR imagery. Nevertheless, precise identification of small-scale non-cultivated habitats faces challenges at both the data foundation and technical methodology levels, which represent a significant bottleneck in the field of agricultural remote sensing.

At the data level, accurate identification of cropland system habitats requires comprehensive consideration of multidimensional factors such as land cover types, spatial locations, morphological characteristics, and vegetation attributes, which imposes extremely high demands on the quality of annotated datasets \citep{tiwariMarginalAgriculturalLand2023}. However, current mainstream remote sensing semantic segmentation datasets (such as Potsdam \citep{songSemanticSegmentationRemoteSensing2020} and Vaihingen \citep{liSemiSupervisedRemoteSensing2023}) are primarily designed for urban scenes and lack high-quality, fine-grained annotations for habitat types in complex agricultural landscapes, making them inadequate for supporting model training and method validation in cropland habitat recognition \citep{songCTMFNetCNNTransformer2023}. Moreover, existing agricultural remote sensing datasets \citep{taoLargeScaleImageText2025} typically focus on crop type mapping or large-scale land cover classification, rarely providing pixel-level annotations for fine-scale non-cultivated habitat elements such as narrow field margins or small ponds. Therefore, In our previous study, we constructed a high-resolution agricultural habitat identification dataset based on the Hailun River Basin, which provided the necessary data foundation for the present research\citep{bianHabitatTypeRecognition2025}.

At the technical level, due to the spatial resolution limitations of medium- and low-resolution remote sensing images, terrain transition zones narrower than 5 m are often difficult to accurately extract because of mixed-pixel effects \citep{yuanReviewDeepLearning2021}. In agricultural landscapes, the spectral mixing between cultivated fields and adjacent non-cultivated elements (e.g., grassy margins or bare ditches) further aggravates this issue, as the narrow spatial extent of these transition zones frequently falls below the sub-pixel scale of medium-resolution sensors \citep{sunMultiSourceRemoteSensing2026}. Traditional remote sensing classification methods typically achieve accuracies below 65\% when dealing with structurally complex or small-scale habitats, falling far short of engineering requirements for fine-grained habitat identification and precise boundary segmentation \citep{liu21stCenturyDaily2021}. The growing availability of VHR remote sensing imagery has stimulated rapid advances in deep learning-based interpretation, which has become the dominant research direction in remote sensing image segmentation \citep{liHybridDualbranchEncoding2026}. Deep learning models such as U-Net \citep{azadMedicalImageSegmentation2024}, DeepLab \citep{chenEncoderDecoderAtrousSeparable2018}, SegFormer \citep{xieSegFormerSimpleEfficient2021}, and DPT (Dense Prediction Transformer) \citep{ranftlVisionTransformersDense2021}, leveraging end-to-end architectures, have significantly improved semantic segmentation performance and effectively enhanced geometric accuracy in field extraction tasks from high-resolution images \citep{suSmallScaleImageSemantic2021}. However, when applied to the extraction of small-scale or fragmented objects in remote sensing imagery---such as narrow roadside edges or scattered abandoned parcels within agricultural matrices---these generic architectures often suffer from incomplete segmentation and boundary ambiguity \citep{hanMultiScaleFeatureLearning2026}. To address the bottleneck of insufficient fusion of multi-level semantic features, Huang et al.\ applied DINOv3 as a feature extraction network within the encoder module, demonstrating that DINOv3 can effectively capture both deep and shallow image features \citep{huangDINOv3GuidedDifference2026}. Dou et al.\ further confirmed the effectiveness of DINOv3 in extracting multi-scale image features for large-scale geological feature recognition; when combined with an MLP decoder, it achieved significant improvements in precision for extracting large-scale contiguous targets while enabling effective integration of multi-layer information \citep{douDINOv3DrivenSemanticSegmentation2026}. Nevertheless, directly transferring this model to non-cropland habitat recognition tasks still presents challenges such as fragmented segmentation of small-scale micro-habitats and blurred boundaries.

In multi-level feature fusion, the way in which the contributions of different layers are combined largely determines the final segmentation quality \citep{dongMultiLevelFeatureFusion2021}. Conventional fusion strategies usually assign fixed, manually determined weights to each level, which cannot adapt to the varying complexity of individual samples and tend to either dilute informative features in complex scenes or overemphasize redundant ones in simple scenes \citep{chenSelfLearningSegmentation2000}. To overcome this limitation, this paper designs a dynamic weighting mechanism in which the fusion weights are adaptively predicted for each sample rather than predefined. More importantly, statistical analysis methods are applied to guide the design of the dynamic weights. We statistically analyzed the fusion weights and their information entropy across the training set: for images containing fewer habitat types, the fusion weights of different layers differ only slightly and exhibit higher weight entropy, whereas for images with more abundant habitat features, the weights diverge considerably and correspond to lower weight entropy. Overall, the weight entropy shows a decreasing trend as the number of habitat categories within an image increases, indicating that the weight distribution becomes more concentrated and stable when facing complex scenes. This reveals that the model can automatically adjust the weights of each layer according to the actual habitat features, concentrating them on more informative layers for complex scenes while exploring and balancing the contributions of different layers more extensively for simple scenes. This statistical insight not only provides a quantitative basis for the dynamic weighting strategy, but also motivates the introduction of an entropy-based regularization term to suppress weight collapse.

To address the aforementioned issues, this paper proposes a Dynamic Weighted Feature Fusion Network (DWFF-Net) tailored for micro-habitat analysis. The main contributions of this paper are as follows:

\begin{enumerate}
  \item We extract multi-level feature information using DINOv3 and employ a lightweight MLP to dynamically predict fusion weights for each layer of features for each sample. This effectively integrates deep semantic information with shallow texture details, maximizing the utilization of multi-level features. An entropy regularization term is also introduced to suppress weight collapse. This approach compensates for the insufficient feature representation of micro-habitats caused by their small scale, thereby significantly improving segmentation accuracy for micro-habitats.
  \item Statistical analysis of the predicted fusion weights and their information entropy reveals that the weight entropy decreases as the number of habitat categories in the scene increases, which quantifies how the network focuses on informative layers in complex scenes.
\end{enumerate}

To our knowledge, this study represents one of the first attempts to apply a dynamically weighted feature fusion strategy driven by DINOv3 to the fine-grained semantic segmentation of micro-habitats in complex agricultural landscapes using VHR remote sensing imagery. By bridging the gap between advanced vision foundation models and domain-specific agricultural remote sensing applications, this work provides a novel methodological paradigm for precise habitat mapping from remotely sensed data.

The remainder of this paper is structured as follows. Section 2 describes the study area and data. The segmentation models are detailed in Section 3. Section 4 introduces the experimental setup, followed by a comparative analysis of the experimental results and an evaluation of the segmentation and recognition quality. Finally, the conclusion and future work are provided in Section 5.

\section{Study area and dataset}
\subsection{Study area}

\begin{figure}[htb]
\centering
\resizebox*{12cm}{!}{\includegraphics{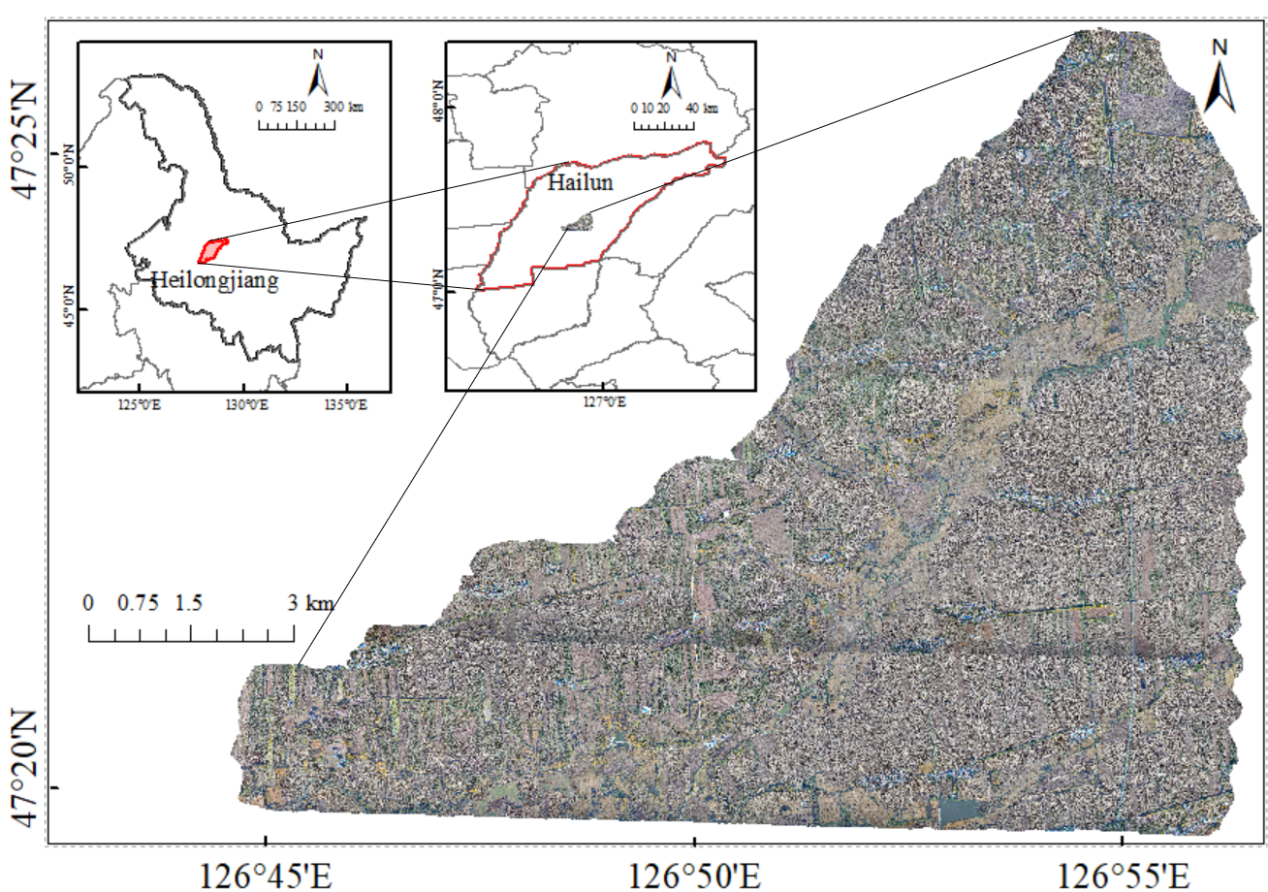}}\hspace{5pt}
\caption{Geographical location and overview of the study area in the Hailun River Basin, Heilongjiang Province, China.} \label{fig1_study_area}
\end{figure}

The experimental area is located in the Hailun River Basin of Hailun City, Heilongjiang Province, China, The experimental area is shown in Figure \ref{fig1_study_area}. spanning from $47^{\circ}19^{\prime}$N to $47^{\circ}27^{\prime}$N and $126^{\circ}44^{\prime}$E to $126^{\circ}57^{\prime}$E. With an average elevation of 201 m and a total area of 74.1 km$^2$, the region is characterized by gently rolling hills (manchuan mangang landform). The primary crops cultivated include maize and soybean. The farmland is often interspersed with shelterbelts, grass strips, and gullies, contributing to a diverse array of habitats within the agricultural landscape.

The ultra-high-resolution remote sensing imagery utilized in this study was acquired by the FeimaRobotics V500 vertical take-off and landing fixed-wing unmanned aerial vehicle (UAV) system. This UAV is equipped with a standard visible spectral sensor (RGB) and a NovAtel high-precision GNSS positioning module. By connecting to a ground-based augmentation network for Real-Time Kinematic (RTK) services, and integrated with the self-developed professional software UAV Manager, the system supports intelligent flight path planning, multi-source data fusion processing, and automated POS data resolution. During image acquisition, the UAV operated at an altitude of 800 m above the small watershed in terrain-following mode, capturing imagery with an ultra-high spatial resolution of 0.08 m.

\subsection{Dataset Creation}

One of the central objectives of the QuESSA (Quantification of Ecological Services for Sustainable Agriculture) project in Europe was to accurately identify the main types of non-cropped habitats within agricultural landscapes and to establish a classification system with broad applicability across the European continent \citep{hollandApproachesIdentifyValue2020}. In this study, sampling was conducted in late September---a period when crops have not yet been harvested and vegetation characteristics are most discernible. During the previous research process \citep{bianHabitatTypeRecognition2025}, following the classification framework developed by the QuESSA project, and based on the key attributes of cropped and non-cropped habitats in the mollisol core region of Northeast China (Hailun City), we developed and validated a habitat dataset applicable to autumn farming systems in this area. The specific habitat types and their principal characteristics are detailed in Table \ref{tab:habitat_classification}.

\begin{table}[htbp]
\centering
\caption{Classification and main characteristics of habitat types in cultivated land system.}
\label{tab:habitat_classification}
\renewcommand{\arraystretch}{1.3}
\resizebox{\textwidth}{!}{%
\begin{tabular}{lllll}
\toprule
Habitat classification & Primary category & Secondary classification & English abbreviation & Main characteristics (end of September) \\
\midrule
\multirow{2}{*}{Cultivated habitats} & \multirow{2}{*}{Cultivated land} & Paddy field & Pf & The field is well-organized; mature rice fields appear golden or yellowish-brown, with clear ridges, minimal standing water, and visible harvesting marks. \\
 &  & Dry land & Dl & Dark yellowish-brown tones with clear ridges; crops mature (yellowish-green speckled) or harvested (bare soil/soil stubble). \\
\midrule
\multirow{13}{*}{Non-cultivated habitats} & \multirow{4}{*}{Forest} & Woody area & Wa & Foliate dark green (evergreen) or yellowish-green (deciduous); high canopy density, rough texture, distinct shadows. \\
 &  & Forest belt & Fb & Stratified dark green (evergreen) or yellowish green (deciduous), with clear borders. \\
 &  & Arbor--Shrub--Grass compound land & Asg & A composite zone of trees, shrubs and herbaceous vegetation; the image tone texture is uneven, dark green and yellow green are interlaced, and the structure is complex. \\
 &  & Scattered trees & St & Scattered individual or small cluster trees in non-forest background, dark green (evergreen) or yellow-green (deciduous) dot or small patch characteristics. \\
\cmidrule(lr){2-5}
 & \multirow{2}{*}{Grass} & Grass belt & Gb & Herbaceous vegetation with elongated stripes; the image appears yellow-green or yellow-brown with uniform texture. \\
 &  & Tidal flats & Tf & The transitional zone between high and low water levels; when the water level is low, the sediment is exposed, and the image appears grayish-white, in strip or patch form. \\
\cmidrule(lr){2-5}
 & \multirow{2}{*}{Water area} & River & River & A linear or narrow strip of water; the image appears dark blue, with yellowish-brown withered grass and grayish-white mudflats visible along the shore. \\
 &  & Water & Water & Stagnant or slow-moving water bodies such as ponds and reservoirs appear dark blue in the image, with visible shoreline vegetation and exposed shallow water forming a tidal flat. \\
\cmidrule(lr){2-5}
 & \multirow{5}{*}{Other} & Paved road & Pr & Artificially paved cement or hardened pavement; images appear bright white or light grey, with regular lines and clear boundaries, no vegetation cover on the pavement. \\
 &  & Dirt road & Dr & The unpaved dirt road is used for agricultural machinery; the image is light brown with a small amount of vegetation and clear wheel tracks. \\
 &  & Construction land & Cl & Village, highway, factory and other artificial construction areas, image tone complex, geometric shape regular, texture rough. \\
 &  & Unused land & Ul & Wasteland (large yellowish brown), saline-alkali land (gray and white), sandy land (bright white), bare land (light brown). \\
 &  & Ridge & Ridge & Narrow earthen ridges used for boundary demarcation and water storage in paddy fields appear light brown or grayish-white, clearly and regularly visible against the field background. \\
\bottomrule
\end{tabular}%
}
\end{table}

\begin{figure}[htb]
\centering
\resizebox*{\textwidth}{!}{\includegraphics{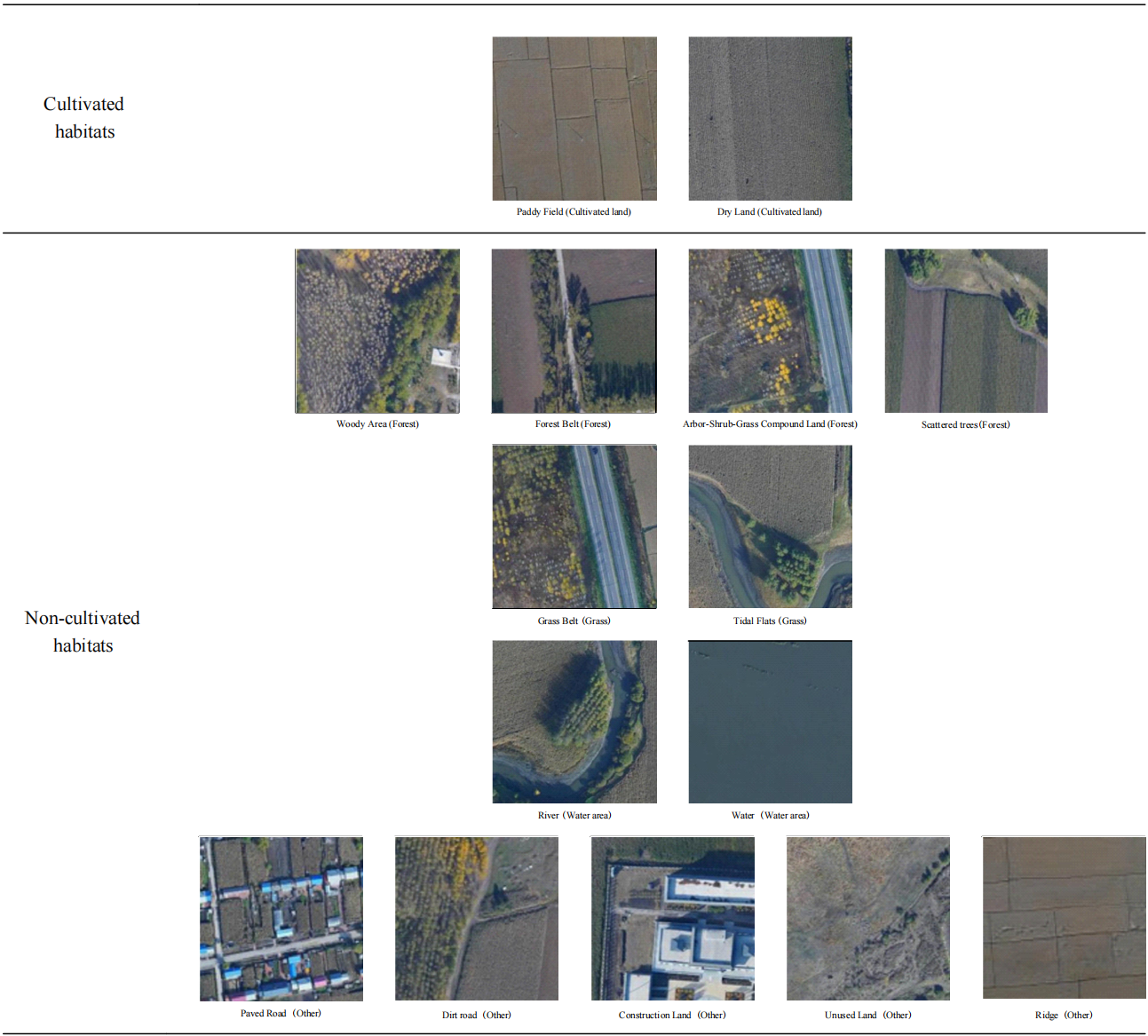}}\hspace{5pt}
\caption{Geographical location and overview of the study area in the Hailun River Basin, Heilongjiang Province, China.} \label{figure2}
\end{figure}

We annotated the images in the dataset according to the classification of the habitat system in Table \ref{tab:habitat_classification}, and provided examples for each habitat category (as shown in Figure \ref{figure2}).

\section{Methodology}

\begin{figure}[htb]
\centering
\resizebox*{\textwidth}{!}{\includegraphics{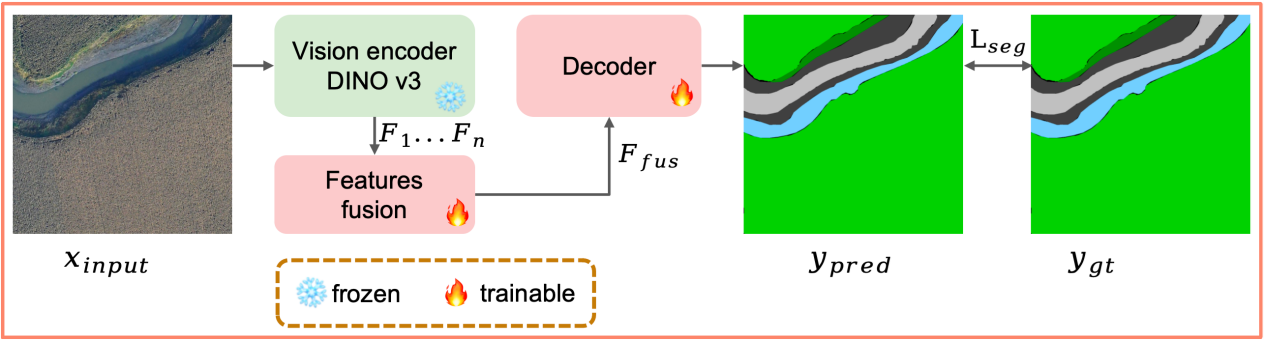}}\hspace{5pt}
\caption{Overall framework of the proposed DWFF-Net.} \label{figure3}
\end{figure}

The overall framework of the proposed cultivated land system habitat model, named DWFF-Net, is illustrated in Figure \ref{figure3}. It comprises the following key components: 1) input image data ${X_{input}}$; 2) a backbone recognition network incorporating a frozen DINOv3 model; 3) the predicted habitat type output ${y_{pred}}$; 4) a supervised training process that compares the prediction with the ground truth label ${y_{gt}}$ using a segmentation loss function ${L_{seg}}$.

\begin{figure}[htb]
\centering
\resizebox*{\textwidth}{!}{\includegraphics{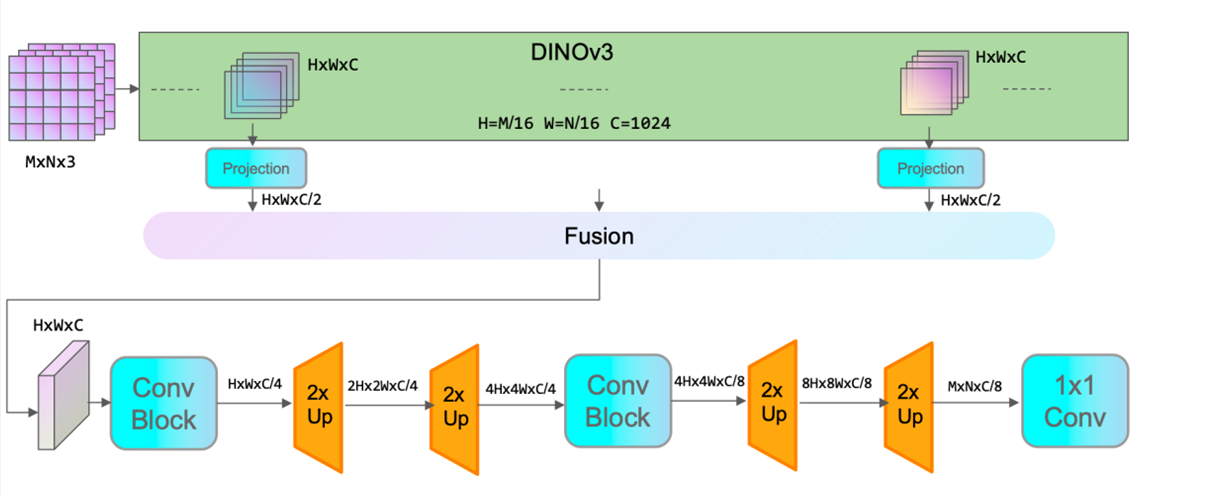}}\hspace{5pt}
\caption{Overall architecture of the proposed DWFF-Net model for~farmland system habitat type identification.} \label{figure4}
\end{figure}

Figure \ref{figure4} depicts the overall architecture of the proposed DWFF-Net. This framework utilizes a frozen DINOv3 encoder as its backbone for feature extraction, integrated with a novel decoder that features a Dynamic Weight Feature Fusion (DWFF) mechanism to produce the final segmentation map. The model is optimized end-to-end via a composite loss function, formulated by integrating an L2-regularized hybrid segmentation loss with the summation of weight entropy across multi-level features.

\subsection{DINOv3 as a Frozen Multi-Level Feature Extractor}

We leverage DINOv3 with a Vision Transformer Large backbone and a patch size of 16 (DINOv3-ViT-L/16) as a frozen multi-level feature extractor. The ViT architecture processes an input image $I \in {R^{H \times W \times 3}}$ by first dividing it into a sequence of non-overlapping patches, which are subsequently linearly projected into patch embeddings. These embeddings are then propagated through a cascade of Transformer blocks to generate hierarchical feature representations.

A fundamental aspect of our approach is that the entire DINOv3 backbone remains frozen throughout the training process. This design offers several key benefits: (1) It substantially reduces the number of trainable parameters, thereby accelerating convergence and lowering memory consumption; (2) It helps prevent overfitting, which is particularly relevant given the limited scale of specialized remote sensing datasets compared to large-scale pre-training data; and (3) It encourages the decoder to learn how to effectively interpret the rich and general-purpose representations from the foundation model, rather than altering its pre-trained feature space.

\subsection{Decoder Design}

The Dynamic-Weighted Feature Fusion Network Decoder (DWFF) is designed to efficiently integrate multi-scale features extracted from the DINOv3 encoder. It aims to combine the precise spatial localization information provided by shallow features with the semantically rich contextual cues captured by deeper layers. The architecture of the decoder comprises three core components: projection, fusion and Dynamic-Weighted Network.

\begin{figure}[htb]
\centering
\resizebox*{\textwidth}{!}{\includegraphics{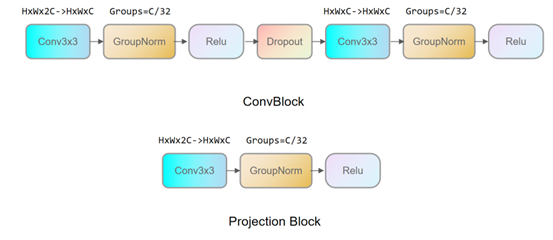}}\hspace{5pt}
\caption{Schematic diagram of the feature projection.} \label{figure5}
\end{figure}

\subsubsection{Feature Projection}

as illustrated in Figure \ref{figure5}, Given that features originating from different layers in a Vision Transformer (ViT) possess identical channel dimensions, we first project each feature map into a shared low-dimensional subspace via a 1$\times$1 convolution. This operation is followed by Group Normalization and ReLU activation function. Such projection not only reduces computational complexity but also facilitates the learning of task-specific feature representations.

\begin{equation}
    {F'_{{l_i}}} = {\rm{ReLU(GN(Con}}{{\rm{v}}_{1 \times 1}}{\rm{(}}{F_{{l_i}}}{\rm{)))}} \in {R^{B \times {C_{fus}} \times {H_p} \times {W_p}}}
\end{equation}

where ${C_{fus}}$ is the fusion channel dimension and $({H_p},{W_p})$ are the patch dimensions of the feature map.

\subsubsection{Dynamic-Weighted Feature Fusion Network (DWFF)}

We initially developed a global weighted feature fusion network, termed Static-Weighted Feature Fusion Network (SWFF-Net), which enhances the utilization of multi-level features to some extent by integrating hierarchical features using predefined fixed weights.

Different feature layers emphasize distinct aspects of habitat information in cultivated land systems, ranging from low-level texture information (Layer 1) to high-level semantic information (Layer 24). However, the simple SWFF-Net model, when fusing these multi-scale features, relies solely on a set of static global weights and fails to adequately account for the dynamic variations in information content across different feature layers under varying input samples. To address this limitation, we further propose a dynamic weighted feature fusion network model, named Dynamic-Weighted Feature Fusion Network (DWFF-Net). By incorporating an input-driven adaptive weight allocation strategy, this model achieves dynamic calibration of the information contained in different feature layers, thereby enabling more effective fusion of multi-level feature information.

\begin{figure}[htb]
\centering
\resizebox*{\textwidth}{!}{\includegraphics{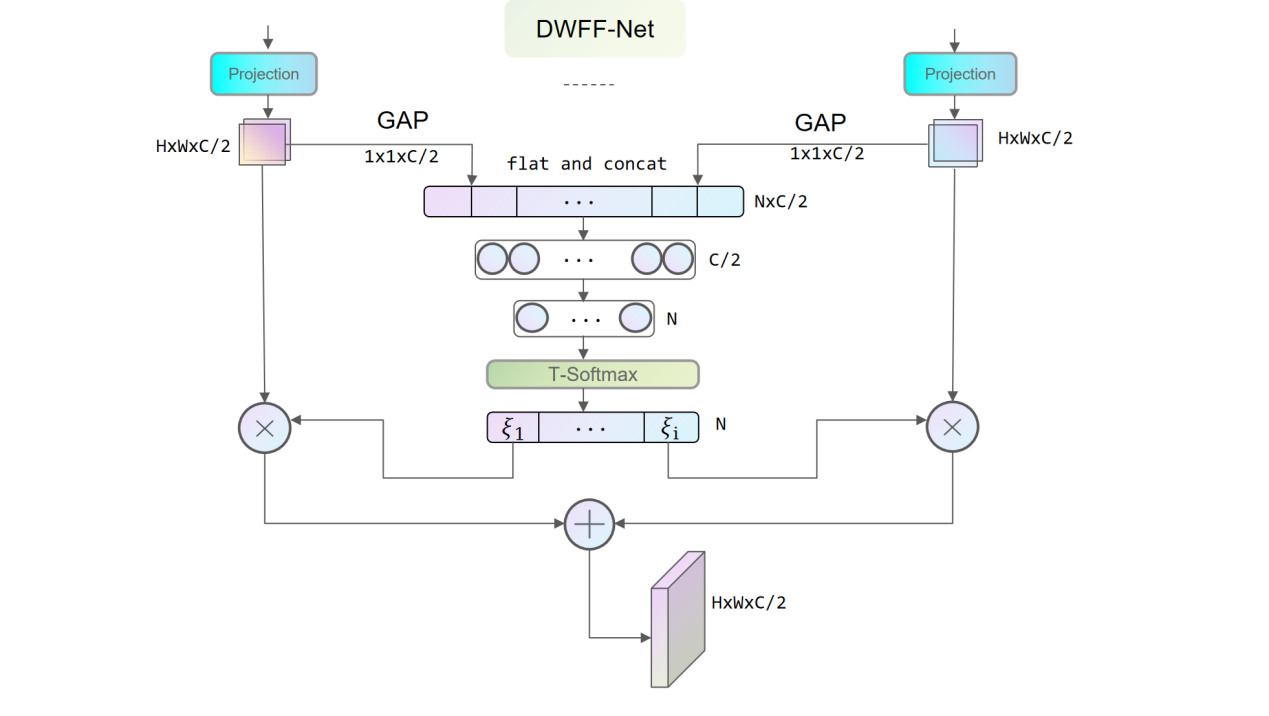}}\hspace{5pt}
\caption{Dynamic-Weighted Feature Fusion (DWFF) mechanism for integrating multi-scale features.} \label{figure6}
\end{figure}

As Figure \ref{figure6}, a learnable weight generation mechanism is employed to adaptively fuse features across levels. Specifically, each level's feature undergoes global average pooling (GAP) to produce a compact vector, which are then concatenated. A two-layer MLP (with ReLU activation) processes the concatenated vector to generate level-wise scores. These scores are normalized via softmax with a learnable temperature parameter, yielding normalized weights for each level.

Consequently, the unified fused feature generated by the proposed dynamic weighting network is expressed as ${F_{fus}}$:
\begin{equation}
    {F_\text{fus}} = \sum\limits_{i = 1}^m {{\xi _i} * {F_i}}
\end{equation}

where $m$ denotes the number of feature layers selected for fusion, and ${\xi _i}$ represents the weight assigned to each layer after Softmax normalization.

\subsection{Overall Loss Function}

As a central component of the supervised learning framework, the loss function not only serves as an objective function guiding parameter updates but also acts as a performance evaluation metric that drives feature learning. In this study, we propose a novel composite loss function architecture:
\begin{equation}
    {L_\text{total}} = {L_\text{seg}} + {\lambda_1}{L_\text{L2}} - {\lambda _2}{L_\text{entropy}}
\end{equation}

Here,$\lambda_1$ and $\lambda_2$ are two adjustment coefficients, which are assigned values of 0.04 and 0.01, respectively, in this study.

To combat the widespread issue of severe class imbalance commonly encountered in agricultural imagery, we implemented a hybrid loss function integrating Dice loss and Focal loss.
\begin{equation}
    {L_\text{seg}} = \alpha {L_\text{Dice}} + \beta {L_\text{Focal}}
\end{equation}

In this study, we design a composite loss function, ${L_\text{seg}}$, which effectively integrates the global region optimization capacity of the Dice loss (${L_\text{Dice}}$) with the hard sample focusing capability of the Focal loss (${L_\text{Focal}}$). The coefficients $\alpha$ and $\beta$ serve as balancing hyperparameters for the respective loss components. This formulation ensures that the segmentation network maintains the overall structural consistency of the target regions while accurately capturing their fine boundary details throughout the training process.

The Dice loss was originally derived from binary tabular data analysis. By enhancing gradient responses for small-scale target structures, the constructed loss function improves regional consistency, making it particularly effective for capturing fine-grained features in habitat classification tasks. The formula for Dice loss is as follows:
\begin{equation}
    {L_\text{Dice}} = \frac{1}{C}\sum\limits_{i = 1}^C {(1 - \frac{{2{{\sum\nolimits_j {\left| {X \cap Y} \right|} }_{i,j}} + \varepsilon }}{{\sum\nolimits_j {{{\left| X \right|}_{i,j}}}  + {{\sum\nolimits_j {\left| Y \right|} }_{i,j}} + \varepsilon }})}
\end{equation}

In this equation, $C$ denotes the total number of habitat categories within the dataset, while $\varepsilon$ represents a smoothing constant incorporated to prevent the denominator of the ${L_\text{Dice}}$ function from becoming 0, especially when negative samples are present. $\left| X \right|$ and $\left| Y \right|$ denote the binary masks of the ground truth and the predicted segmentation, respectively.

The Focal loss function effectively emphasizes hard examples, such as those in shadowed regions, by automatically assigning higher weights to low-confidence samples, thereby mitigating class imbalance. The formulation of the Focal loss is provided as follows:
\begin{equation}
    {L_\text{Focal}} =  - {\alpha _t}{(1 - {p_t})^\gamma }\log ({p_t})
\end{equation}

In the equation, $p_{t}$ denotes the predicted probability of the target class by the model; $\alpha_{t}$ is a balance factor that adjusts the influence between positive and negative samples; and $\gamma$ is the focal factor, which modulates the weight assigned to hard and easy examples, thus improving the focus of the model on challenging cases and improving its overall calibration.

The introduction of the L2 regularization term $L_\text{L2}$ enhances the robustness of the optimization process, particularly in high-dimensional scenarios where overfitting is frequently encountered. By discouraging excessively large weight values, this regularization term promotes model simplicity, improves generalization capability, and reduces variance without substantially increasing bias. The formulation of $L_\text{L2}$ is given as follows:
\begin{equation}
    {L_\text{L2}} = \sum\nolimits_i {\omega _i^2}
\end{equation}

In the equation, $\omega_{i}$ denotes an individual weight parameter of the model.

To ensure the accuracy of multi-layer fusion and prevent weight collapse, the total entropy of each feature layer in the DINOv3 network must be calculated, denoted as $L_\text{entropy}$ , is introduced. Entropy serves as a metric for the uniformity of a probability distribution: a higher entropy value indicates a more uniform weight distribution (e.g., when all weights are 0.25, entropy reaches its maximum), whereas a lower entropy value suggests a more concentrated distribution (e.g., when one weight approaches 1 while others are close to 0). By subtracting the $L_\text{entropy}$ term from the total loss function, the aim is to mitigate the issue of excessively concentrated weight distributions. Since the loss function is optimized toward minimization, a smaller entropy value (which contributes more due to the negative sign) leads to a larger total loss and consequently generates stronger gradients. During gradient descent, this drives the model to adjust its parameters in a way that increases entropy, thereby promoting a more uniform weight distribution.

The selection of feature layers is designed to capture multi-scale representations, spanning from low-level textures (Layer 1) to high-level semantic information (Layer 24). Taking into account both computational complexity and experimental constraints, four specific layers---namely, Layers 1, 8, 16, and 24---were systematically sampled, and the sum of their information entropy was computed. The formulation of the entropy loss term, $L_\text{entropy}$ , is given as follows:
\begin{equation}
    {L_\text{entropy}} =  - \sum {{p_i}(x)\log {p_i}(x)}
\end{equation}

In the equation, ${p_i}(x)$ represents the probability distribution derived from the normalized feature activations of a specific layer.

\subsection{Weight collapse}

In Section 3.2, we developed the DWFF-Net architecture, which employs a data-driven strategy to dynamically generate weights for feature fusion. During the training process, however, the attention mechanism may disproportionately and persistently favor a narrow subset of features---such as those from specific layers or channels---leading to a phenomenon termed "attention dictatorship." This issue induces weight collapse, wherein a limited number of features dominate the attention distribution, while the contributions of other potentially informative features are significantly suppressed. Consequently, the performance gains that the attention mechanism is intended to deliver are undermined.

The $L_\text{entropy}$ introduced in Section 3.3 partially addresses this phenomenon, and further mitigation is required. a temperature hyperparameter ($temp$) is introduced to modulate the original attention scores. Specifically, the final attention weights are computed by applying a softmax function to the temperature-scaled $scores$:
\begin{equation}
    Weight = {F_{{\rm{softmax}}}}(\frac{{scores}}{{temp}})
\end{equation}

This temperature parameter functions as a smoothing factor for probability distribution, effectively controlling the degree of weight distribution smoothing and entropy. When the temperature is high ( $temp$ > 1), the weight distribution becomes more uniform, enhancing the model's ability to explore different feature layers. Conversely, low-temperature settings ( $temp$ < 1) amplify the weights of highly scored features, resulting in a more concentrated distribution. By appropriately adjusting the temperature, we can mitigate attentional bias caused by weight distribution polarization, facilitate effective fusion of multi-source features, and improve the model's representational capabilities and generalization performance.

\subsection{Model Performance Evaluation}
The evaluation of semantic segmentation algorithms generally adheres to well-established conventions in the computer vision field, utilizing metrics such as Precision, Recall, F1-score, and Intersection-over-Union (IoU). In binary classification tasks, each pixel is typically categorized into one of four classes: True Positive (TP), True Negative (TN), False Positive (FP), and False Negative (FN). The relevant performance metrics are defined as follows:
\begin{gather}
    Precision = \frac{{TP}}{{TP + FP}}\tag{10a}\\
    Recall = \frac{{TP}}{{TP + FN}}\tag{10b}\\
    F1 = \frac{{2 \times Precision \times Recall}}{{Precision + Recall}}\tag{10c}\\
    IoU = \frac{{TP}}{{TP + FP + FN}}\tag{10d}
\end{gather}

To comprehensively evaluate and compare the overall performance of different models, we introduce the following metrics: mean Precision (mPrecision), mean Recall (mRecall), mean F1-score (mF1), and mean Intersection over Union (mIoU). The corresponding formulas are provided below.
\begin{gather}
    mPrecision = \frac{{\sum\limits_{i = 1}^{15} {Precisio{n_i}} }}{{15}}\tag{11a}\\
    mRecall = \frac{{\sum\limits_{i = 1}^{15} {Recal{l_i}} }}{{15}}\tag{11b}\\
    mF1 = \frac{{\sum\limits_{i = 1}^{15} F {1_i}}}{{15}}\tag{11c}\\
    mIoU = \frac{{\sum\limits_{i = 1}^{15} {Io{U_i}} }}{{15}}\tag{11d}
\end{gather}

\section{Experiments and result}

\subsection{Experimental details and Training Parameters}

Our proposed framework was implemented based on PyTorch and PyTorch Lightning. Training was conducted using a batch size of 4 ,However, based on this, gradient accumulation was performed every 8 rounds, resulting in an effective batch size of 32, distributed on two RTX 2080Ti 11GB GPUs. To accelerate training, we leveraged mixed-precision training (FP16 mixed-precision) throughout the experiments.

During the training process, we utilized the pre-trained DINOv3-ViT-L/16 model with frozen weights. We divided the dataset of 800 total records into train, val, and test sets in a 6:1:1 ratio. For the DWFF-Net, we extracted features from Transformer blocks {1, 8, 16, 24}. The fusion channel dimension $C_{fus}$ was set to 512. The patch dimensions of the feature map $(H_p,W_p)$ was set to (1248,1248).All models were trained for 150 epochs using the AdamW optimizer with a cosine-annealed learning rate. Standard augmentations including random cropping, flipping, and rotation were employed.

\subsection{Weighted Entropy Research}

\begin{figure}[htb]
\centering
\resizebox*{\textwidth}{!}{\includegraphics{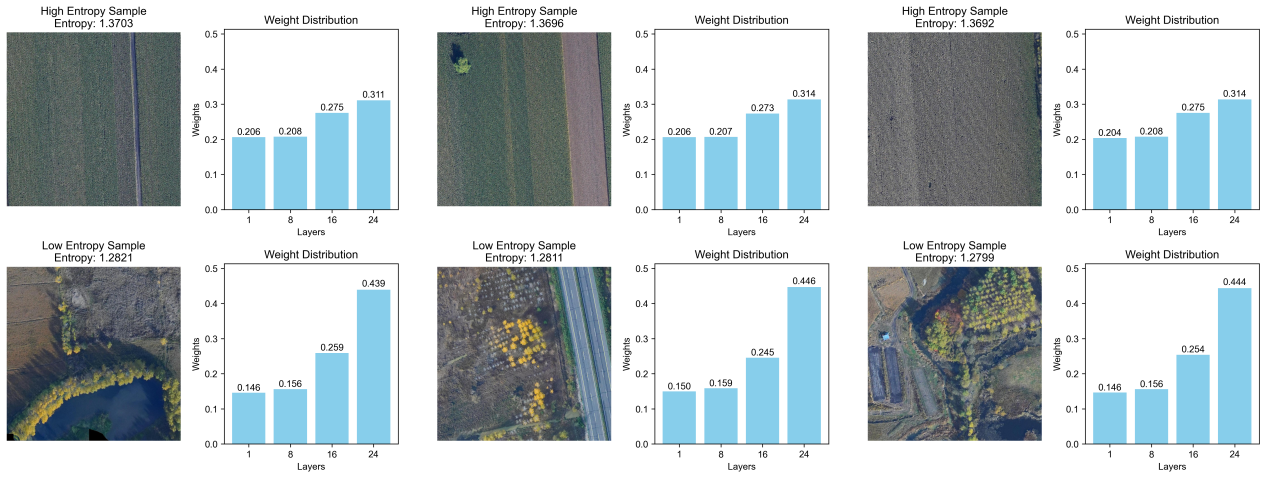}}\hspace{5pt}
\caption{Histograms of layer-wise weights and corresponding entropy values for high-entropy and low-entropy images.} \label{figure7}
\end{figure}

We investigated the relationship between the weight entropy and the number of habitat categories present in the images. We selected three high - entropy images and three low-entropy images, extracted their corresponding weights from the multi-layer fusion module of the DWFF - Net model, and plotted the weight histograms of different layers while indicating the weight entropy values corresponding to the images. As shown in the first three images in Figure \ref{figure7}, for images containing fewer habitat types, the weight parameters during the fusion of each layer have relatively small differences and exhibit higher weight entropy values. Conversely, the last three images in Figure \ref{figure7} indicate that for images with more abundant habitat feature information, the weight parameters during the fusion of each layer have relatively large differences and correspond to lower weight entropy values.

\begin{figure}[htb]
\centering
\resizebox*{10cm}{!}{\includegraphics{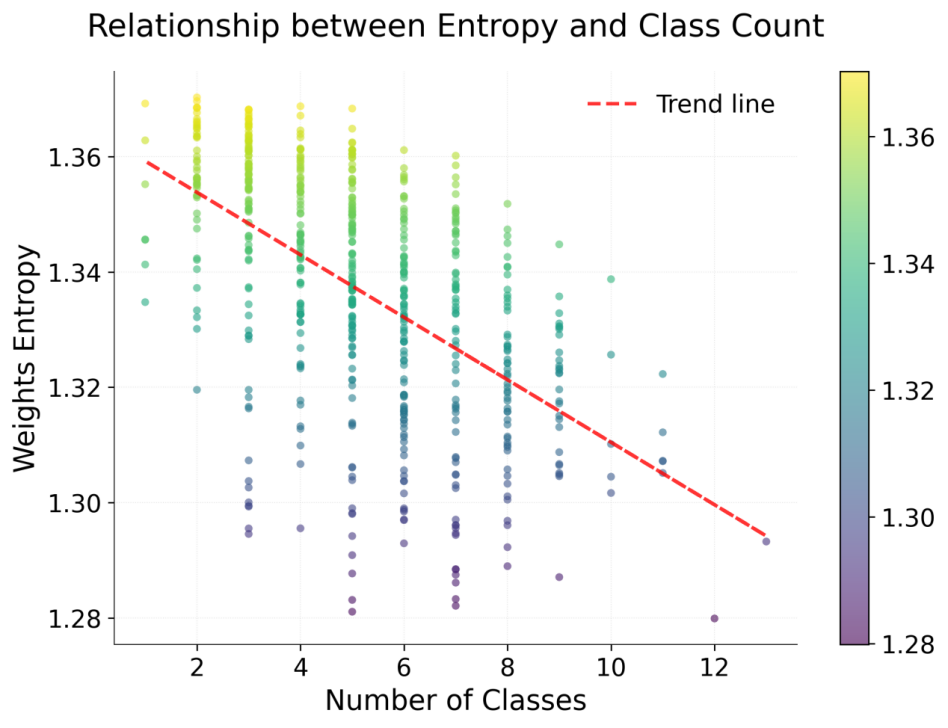}}\hspace{5pt}
\caption{Relationship between weight entropy and the number of habitat categories present in images.} \label{figure8}
\end{figure}

Through statistical analysis of the fusion weights and weight entropy values of these six images, we initially found that the value of the image's weight entropy decreases as the number of features contained in the image increases. Subsequently, we calculated the weight entropy values corresponding to each image in the training set and plotted the relationship diagram between the weight entropy and the number of habitat categories present in the images, as shown in Figure \ref{figure8}.

As shown in Figure \ref{figure8}, as the number of habitat features in the image increases, the weight entropy shows a downward trend. This phenomenon indicates that when the image contains more diverse habitat categories and richer feature information, the weight distribution among the layers in the multi-layer fusion module of the model becomes more concentrated and stable. From a data-driven perspective, this means that the model can automatically adjust the weights of each layer according to the actual number of habitat features in the image, thus more effectively capturing and integrating key information. When the number of habitat features is limited, the model may need to more extensively explore and balance the contributions of different layers in the multi-layer fusion module, which will result in a more dispersed weight distribution and consequently a higher weight entropy value. In contrast, as the number of habitat features increases, the model will be more capable of identifying key features and assigning larger weights to relevant layers, thereby achieving a more focused weight distribution and correspondingly reducing the weight entropy value.

We speculate that the relationship between weight entropy and the number of habitat categories may reflect the adaptability of the model when dealing with data of different complexities. In practical applications, this characteristic enables the model to more effectively utilize feature information from different levels when facing complex and diverse habitat scenarios, thus improving the accuracy of habitat type recognition.

\begin{figure}[htb]
\centering
\resizebox*{10cm}{!}{\includegraphics{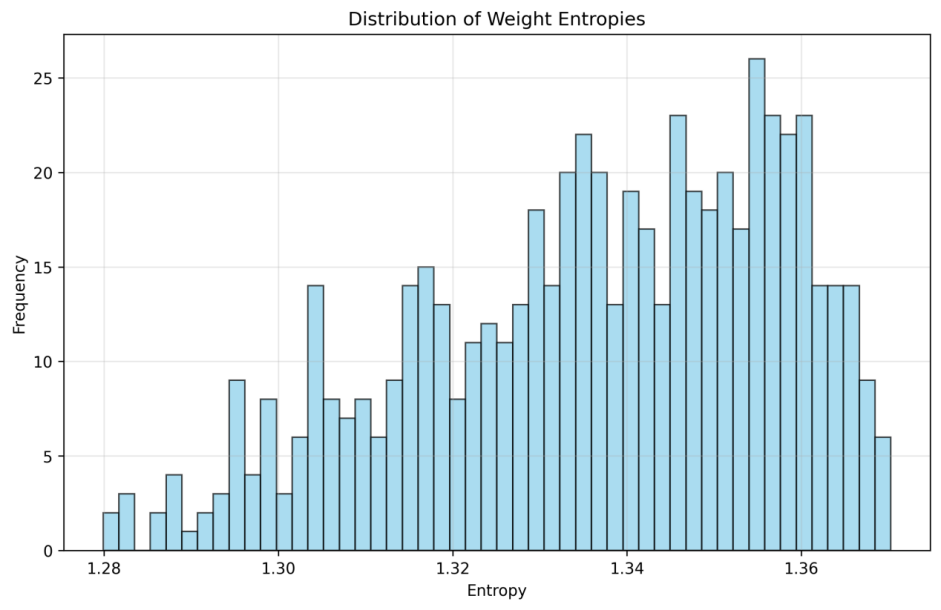}}\hspace{5pt}
\caption{Histogram of the overall distribution of weight entropy values.} \label{figure9}
\end{figure}

Furthermore, we also visualized the overall distribution of the weight entropy through a histogram. The results are shown in Figure \ref{figure9}: In the high entropy region, the distribution range is more extensive; while in the low entropy region, the distribution is more concentrated. This pattern indicates that the model's ability to handle different complexity of habitat features varies. The wide distribution in the high entropy region may reflect that when processing images with relatively homogeneous and less informative habitat features, the model requires more exploration and adaptation processes, resulting in increased uncertainty in the weights and a more dispersed entropy distribution; conversely, the concentrated distribution in the low entropy region indicates that when the image contains diverse and rich habitat features, the model can determine the layer weights more efficiently, thereby generating a relatively stable and concentrated entropy distribution of weights.

Through in-depth study of the weight entropy distribution, we not only clarified the adaptive mechanism of this model when dealing with data from different complexity habitats, but also laid an important foundation for subsequent model optimization. In future research, targeted improvements can be made based on the characteristics of the weight entropy distribution - for example, adjusting the structure of the multi-layer fusion module or optimizing the weight allocation strategy, thereby enhancing the model's recognition performance in diverse habitat scenarios.

\subsection{Qualitative Results}

To rigorously validate the contribution of each component in the proposed framework, we conducted comprehensive ablative experiments. The baseline model employs a non-weighted multi-layer feature fusion network, termed Non-weighted Feature Fusion Network with L layers (NWFF-Net-L), where L denotes the number of fused layers. For comparison, a Static Weighted Feature Fusion Network (SWFF-Net) was also established to verify the effectiveness of dynamic weight-based fusion. Throughout the ablation studies, the core innovation of this work---a multi-layer feature fusion network with dynamic weighting---was progressively integrated into the model. All models were trained and evaluated on the same dataset under identical experimental configurations, including learning rate, optimizer, and number of training epochs. Quantitative results are detailed in Table \ref{tab:ablation_train}.

\begin{figure}[htb]
\centering
\resizebox*{12cm}{!}{\includegraphics{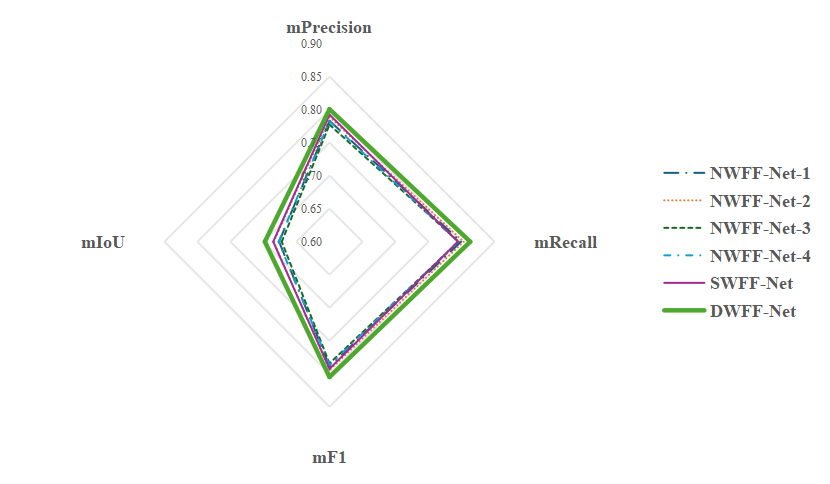}}\hspace{5pt}
\caption{Comparison of training accuracy among the DWFF-Net, SWFF-Net, and NWFF-Net-L models.} \label{figure10}
\end{figure}

As illustrated in Table \ref{tab:ablation_train}, the baseline model Non-weighted Feature Fusion Network (single layer) (NWFF-Net-1) has already demonstrated outstanding performance, fully illustrating the advantage of the DINOv3 feature extraction technique. As illustrated in Figure \ref{figure10}, With the introduction of a multi-level feature fusion mechanism incorporating global weighting, the SWFF-Net model achieved a notable improvement in overall mIoU compared to the standard model, with an increase of 1.3\%. Furthermore, by adopting a dynamic weighting strategy in the multi-level feature fusion mechanism, the DWFF-Net model exhibited an even more significant enhancement in overall mIoU over the SWFF-Net, with an increase of 1.8\%. More importantly, the IoU for the Scattered trees (St) also increased markedly by nine percentage points, further validating that the dynamic integration of shallow texture features and deep semantic features plays a crucial role in fully leveraging multi-level feature information.

\begin{table}[htbp]
\centering
\caption{Ablation study results comparing different feature fusion strategies.}
\label{tab:ablation_train}
\renewcommand{\arraystretch}{1.2}
\begin{tabular}{lcccc}
\toprule
Method & mPrecision & mRecall & mF1 & mIoU \\
\midrule
NWFF-Net-1 & 0.7831 & 0.7992 & 0.7888 & 0.6763 \\
NWFF-Net-2 & 0.7888 & 0.8032 & 0.7946 & 0.6844 \\
NWFF-Net-3 & 0.7772 & 0.7956 & 0.7843 & 0.6723 \\
NWFF-Net-4 & 0.7822 & 0.7946 & 0.7870 & 0.6771 \\
SWFF-Net   & 0.7925 & 0.7944 & 0.7925 & 0.6852 \\
\textbf{DWFF-Net} & \textbf{0.8006} & \textbf{0.8131} & \textbf{0.8049} & \textbf{0.6979} \\
\bottomrule
\end{tabular}
\end{table}

\begin{figure}[htb]
\centering
\resizebox*{12cm}{!}{\includegraphics{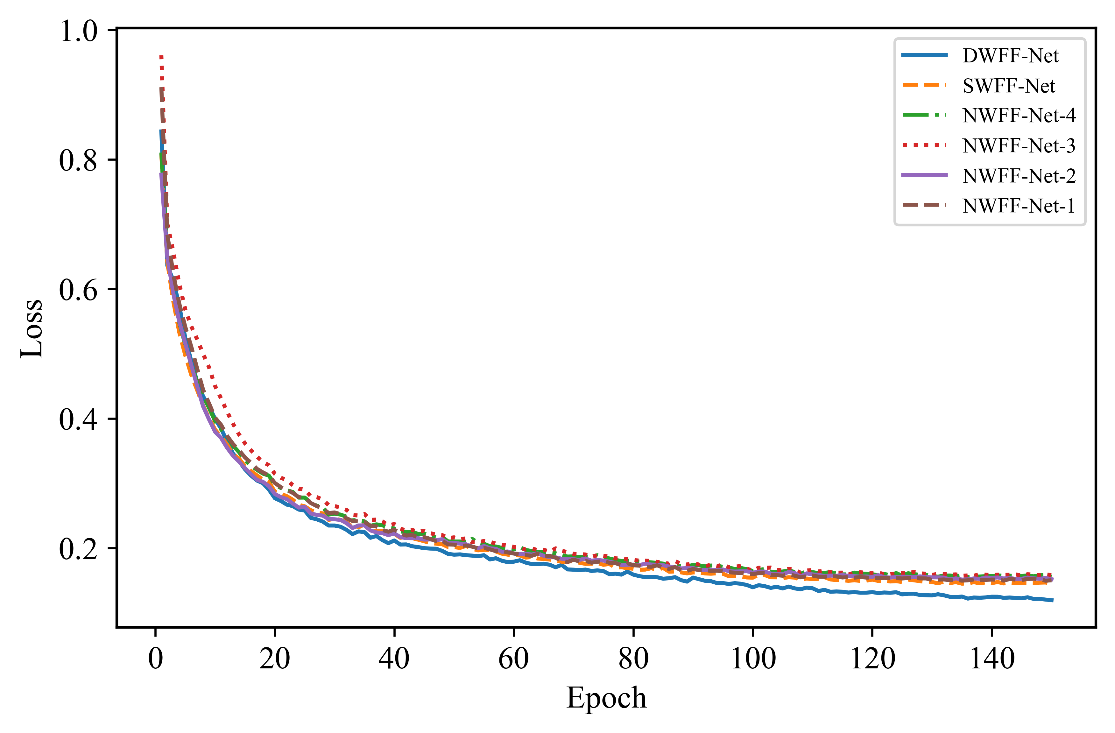}}\hspace{5pt}
\caption{Training loss curves for the DWFF-Net, SWFF-Net, and NWFF-Net-L models.} \label{figure11}
\end{figure}

Experimental results with the hybrid loss function (as illustrated in Figure \ref{figure11}) indicate that the DWFF-Net model demonstrates significant performance advantages throughout 150 training iterations. Its final training loss (0.114) was lower than that of the baseline model (0.156), confirming the effectiveness of multi-level feature fusion. Additionally, after 80 iterations, the rate of loss reduction for the DWFF-Net model surpassed that of the SWFF-Net, underscoring the efficacy of the proposed dynamic weighting network in multi-level feature fusion.

\begin{figure}[htb]
\centering
\resizebox*{12cm}{!}{\includegraphics{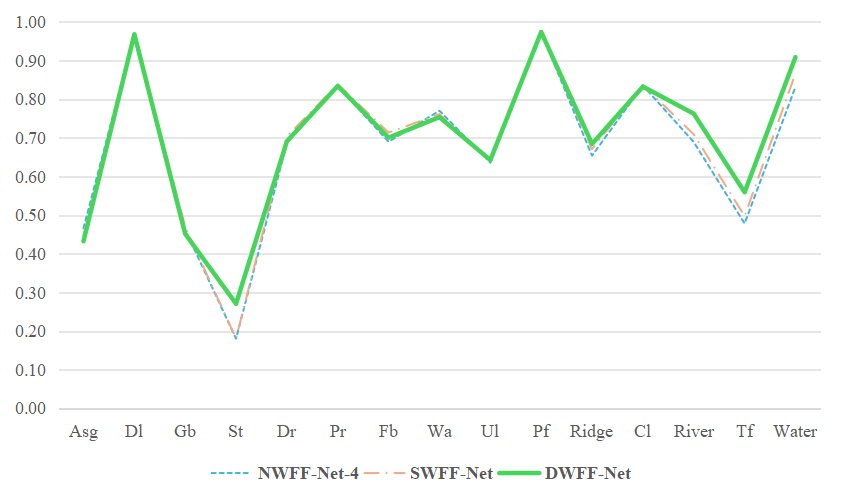}}\hspace{5pt}
\caption{Comparison of segmentation accuracy using different fusion strategies.} \label{figure12}
\end{figure}

The performance of different models NWFF-Net-L, SWFF-Net and DWFF-Net in habitat recognition is summarized in Table \ref{tab:ablation_perclass}. Combining the statistical data in Figure \ref{figure12} and Table \ref{tab:ablation_perclass}, specifically: The DWFF-Net model with dynamic weight feature image fusion achieved IoU values of 0.2707, 0.5594 and 0.7626 on Scattered trees (St), Tidal flats (Tf) and River respectively, which were 33.36\%, 14.48\% and 9.62\% higher than the non-weighted fusion method of NWFF-Net-4, and 33.28\%, 11.8\% and 7.00\% higher than the static weight fusion recognition method of SWFF-Net. In conclusion, the model after multi-layer feature dynamic fusion significantly improves the extraction accuracy of micro-scale habitat features.

\begin{figure}[htb]
\centering
\resizebox*{\textwidth}{!}{\includegraphics{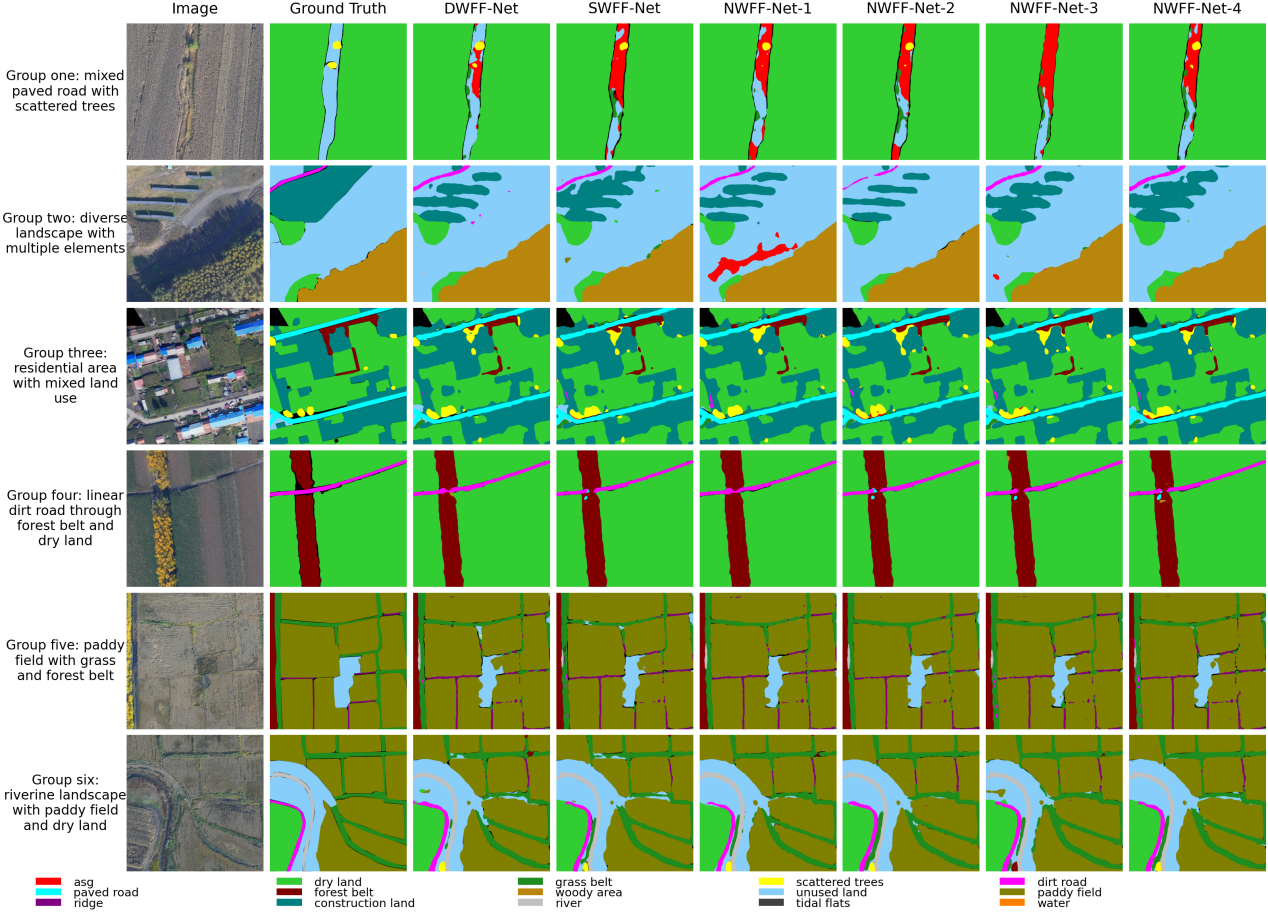}}\hspace{5pt}
\caption{Comparison of extraction effects of DWFF-Net, SWFF-Net and different NWFF-Net.} \label{figure13}
\end{figure}

As shown in Figure \ref{figure13}, the visual recognition results of the three models NWFF-Net-4, SWFF-Net and DWFF-Net on typical habitat images are compared. It can be intuitively seen that NWFF-Net-4, due to not weighting and distinguishing different hierarchical features, has more false detections and omissions for small area habitat patches, and the edge contour recognition results are also relatively blurry; SWFF-Net, after adopting static weight fusion, has improved the accuracy of habitat area division, but still has limited ability to distinguish features in complex scenes, and still has some problems of discontinuous segmentation of similar habitats; while the DWFF-Net proposed in this paper, with the dynamic weighted multi-layer feature fusion strategy, can accurately capture different scales of habitat features, not only complete the identification of different types of habitat areas, but also the segmentation edges are closer to the real annotation, and the recognition accuracy of small area scattered habitats is particularly improved, further intuitively verifying the effectiveness of the method in this paper.

\begin{table}[htbp]
\centering
\caption{F1 and IoU values of 15 habitat types in different feature fusion strategies.}
\label{tab:ablation_perclass}
\renewcommand{\arraystretch}{1.2}
\resizebox{\textwidth}{!}{%
\begin{tabular}{lcccccccccccc}
\toprule
Class & \multicolumn{2}{c}{NWFF-Net-1} & \multicolumn{2}{c}{NWFF-Net-2} & \multicolumn{2}{c}{NWFF-Net-3} & \multicolumn{2}{c}{NWFF-Net-4} & \multicolumn{2}{c}{SWFF-Net} & \multicolumn{2}{c}{DWFF-Net} \\
\cmidrule(lr){2-3} \cmidrule(lr){4-5} \cmidrule(lr){6-7} \cmidrule(lr){8-9} \cmidrule(lr){10-11} \cmidrule(lr){12-13}
 & F1 & IoU & F1 & IoU & F1 & IoU & F1 & IoU & F1 & IoU & F1 & IoU \\
\midrule
\textbf{Asg}   & 0.6005 & 0.4291 & 0.6273 & 0.4570 & 0.6246 & 0.4541 & 0.6358 & 0.4660 & 0.6181 & 0.4473 & \textbf{0.6040} & \textbf{0.4326} \\
\textbf{Dl}    & 0.9831 & 0.9667 & 0.9833 & 0.9671 & 0.9838 & 0.9681 & 0.9842 & 0.9689 & 0.9841 & 0.9687 & \textbf{0.9839} & \textbf{0.9682} \\
\textbf{Gb}    & 0.6426 & 0.4734 & 0.6457 & 0.4768 & 0.6570 & 0.4892 & 0.6322 & 0.4622 & 0.6391 & 0.4696 & \textbf{0.6222} & \textbf{0.4516} \\
\textbf{St}    & 0.3604 & 0.2198 & 0.3781 & 0.2331 & 0.3189 & 0.1897 & 0.3057 & 0.1804 & 0.3059 & 0.1806 & \textbf{0.4261} & \textbf{0.2707} \\
\textbf{Dr}    & 0.8243 & 0.7011 & 0.8233 & 0.6997 & 0.8235 & 0.6999 & 0.8145 & 0.6871 & 0.8262 & 0.7038 & \textbf{0.8175} & \textbf{0.6913} \\
\textbf{Pr}    & 0.9071 & 0.8300 & 0.9075 & 0.8306 & 0.9182 & 0.8487 & 0.9092 & 0.8336 & 0.9121 & 0.8384 & \textbf{0.9098} & \textbf{0.8345} \\
\textbf{Fb}    & 0.8261 & 0.7038 & 0.8235 & 0.6999 & 0.8142 & 0.6866 & 0.8166 & 0.6900 & 0.8330 & 0.7138 & \textbf{0.8241} & \textbf{0.7008} \\
\textbf{Wa}    & 0.8639 & 0.7605 & 0.8635 & 0.7597 & 0.8726 & 0.7741 & 0.8703 & 0.7703 & 0.8647 & 0.7616 & \textbf{0.8595} & \textbf{0.7537} \\
\textbf{Ul}    & 0.7727 & 0.6296 & 0.7831 & 0.6436 & 0.7787 & 0.6376 & 0.7774 & 0.6358 & 0.7807 & 0.6403 & \textbf{0.7825} & \textbf{0.6427} \\
\textbf{Pf}    & 0.9854 & 0.9711 & 0.9861 & 0.9726 & 0.9858 & 0.9721 & 0.9863 & 0.9730 & 0.9865 & 0.9733 & \textbf{0.9869} & \textbf{0.9741} \\
\textbf{Ridge} & 0.7908 & 0.6540 & 0.7978 & 0.6637 & 0.7890 & 0.6515 & 0.7908 & 0.6540 & 0.8029 & 0.6707 & \textbf{0.8127} & \textbf{0.6846} \\
\textbf{Cl}    & 0.9061 & 0.8283 & 0.9064 & 0.8288 & 0.9086 & 0.8325 & 0.9105 & 0.8357 & 0.9112 & 0.8368 & \textbf{0.9088} & \textbf{0.8328} \\
\textbf{River} & 0.7713 & 0.6278 & 0.8186 & 0.6928 & 0.7615 & 0.6149 & 0.8160 & 0.6892 & 0.8299 & 0.7092 & \textbf{0.8653} & \textbf{0.7626} \\
\textbf{Tf}    & 0.7069 & 0.5467 & 0.6541 & 0.4860 & 0.6470 & 0.4782 & 0.6472 & 0.4784 & 0.6643 & 0.4974 & \textbf{0.7175} & \textbf{0.5594} \\
\textbf{Water} & 0.8906 & 0.8027 & 0.9214 & 0.8543 & 0.8813 & 0.7877 & 0.9083 & 0.8320 & 0.9287 & 0.8668 & \textbf{0.9523} & \textbf{0.9090} \\
\bottomrule
\end{tabular}%
}
\end{table}

\subsection{Comparison Experiment}

Finally, we conducted comparative experiments on the training and recognition effects of the DWFF-Net model proposed in this paper with several mature segmentation models. To ensure the fairness of the experiments, all models were pre-trained on our dataset using ImageNet.

\begin{figure}[htb]
\centering
\resizebox*{12cm}{!}{\includegraphics{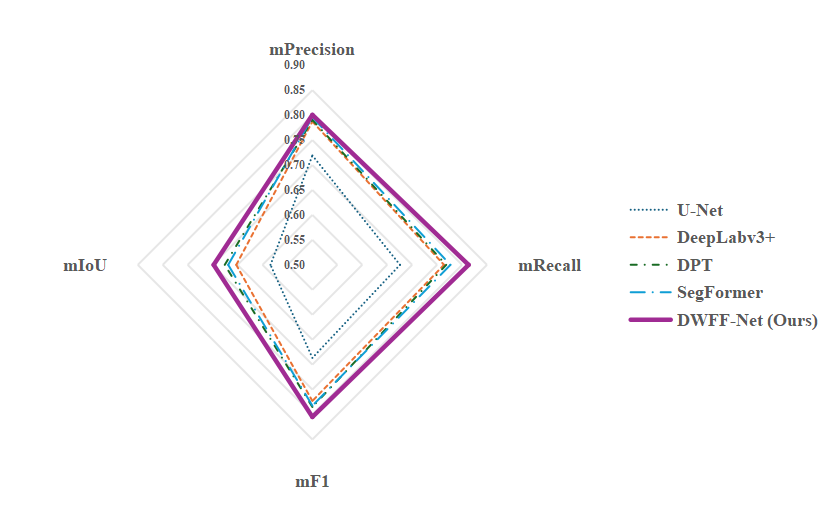}}\hspace{5pt}
\caption{Comparative evaluation of training accuracy across different segmentation models.} \label{figure14}
\end{figure}

As shown in Figure \ref{figure14}, the DWFF-Net model proposed in this paper significantly outperforms the other comparison models in multiple performance indicators, with a prominent performance advantage. According to the specific statistical results in Table \ref{tab:model_comparison_train}, in the mF1 indicator, the DWFF-Net model is 0.8049, which is 14.71\% higher than the worst-performing U-Net model and 2.48\% higher than the second-best DPT model; in the mIoU indicator, the DWFF-Net model is 0.6979, which is 16.32\% higher than the worst-performing U-Net model and 3.18\% higher than the second-best DPT model. The above experimental results prove that the method combining DINOv3 and dynamic weight feature fusion proposed in this paper has excellent performance in handling complex image segmentation tasks with multiple features.

\begin{table}[htbp]
\centering
\caption{Performance comparison of DWFF-Net with other segmentation models.}
\label{tab:model_comparison_train}
\renewcommand{\arraystretch}{1.2}
\resizebox{\textwidth}{!}{%
\begin{tabular}{llcccc}
\toprule
Model & Backbone & mPrecision & mRecall & mF1 & mIoU \\
\midrule
U-Net     & ResNet-50       & 0.7192 & 0.6756 & 0.6865 & 0.5840 \\
DeepLabv3+ & ResNet-50      & 0.7873 & 0.7638 & 0.7729 & 0.6526 \\
DPT       & ViT             & 0.7905 & 0.7682 & 0.7849 & 0.6757 \\
SegFormer & Mit-b2          & 0.7960 & 0.7767 & 0.7824 & 0.6688 \\
\textbf{DWFF-Net (Ours)} & \textbf{Frozen DINOv3-L} & \textbf{0.8006} & \textbf{0.8131} & \textbf{0.8049} & \textbf{0.6979} \\
\bottomrule
\end{tabular}%
}
\end{table}

Table \ref{tab:iou_comparison} summarizes the IoU values of the U-Net, DeepLabV3+, DPT, SegFormer, and DWFF-Net models for the segmentation and recognition of 15 habitat types. Specifically, in the segmentation of large-scale cultivated land habitats, the IoU value of Paddy field (Pf) is 0.9714, which is the smallest difference compared to the worst-performing DeepLabv3+ model (0.9414) and the second-best DPT model (0.9663); the IoU value of Dry land (Dl) is 0.9682, with the smallest difference compared to the worst-performing U-Net model (0.9521) and the second-best SegFormer model (0.9648); in the segmentation of small-scale non-cultivated land habitats, the IoU value of Scattered trees (St) is 0.2707, which is 99.85\% higher than the worst-performing U-Net model and 11.45\% higher than the second-best DeepLabv3+ model; the IoU value of Forest belt (Fb) is 0.7008, with a 12.79\% increase compared to the worst-performing U-Net model and a 6.28\% increase compared to the second-best SegFormer model; the IoU value of Grass belt (Gb) is 0.4516, with a 13.95\% increase compared to the worst-performing U-Net model and a 1.08\% increase compared to the second-best DeepLabv3+ model. In conclusion, the DWFF-Net model proposed in this paper maintains the segmentation accuracy of large-scale cultivated land habitats (e.g. Paddy field, Dry land) while improving the segmentation accuracy of small-scale non-cultivated land habitats (e.g. Scattered trees, Forest belt, Grass belt) to a certain extent.

\begin{figure}[htb]
\centering
\resizebox*{\textwidth}{!}{\includegraphics{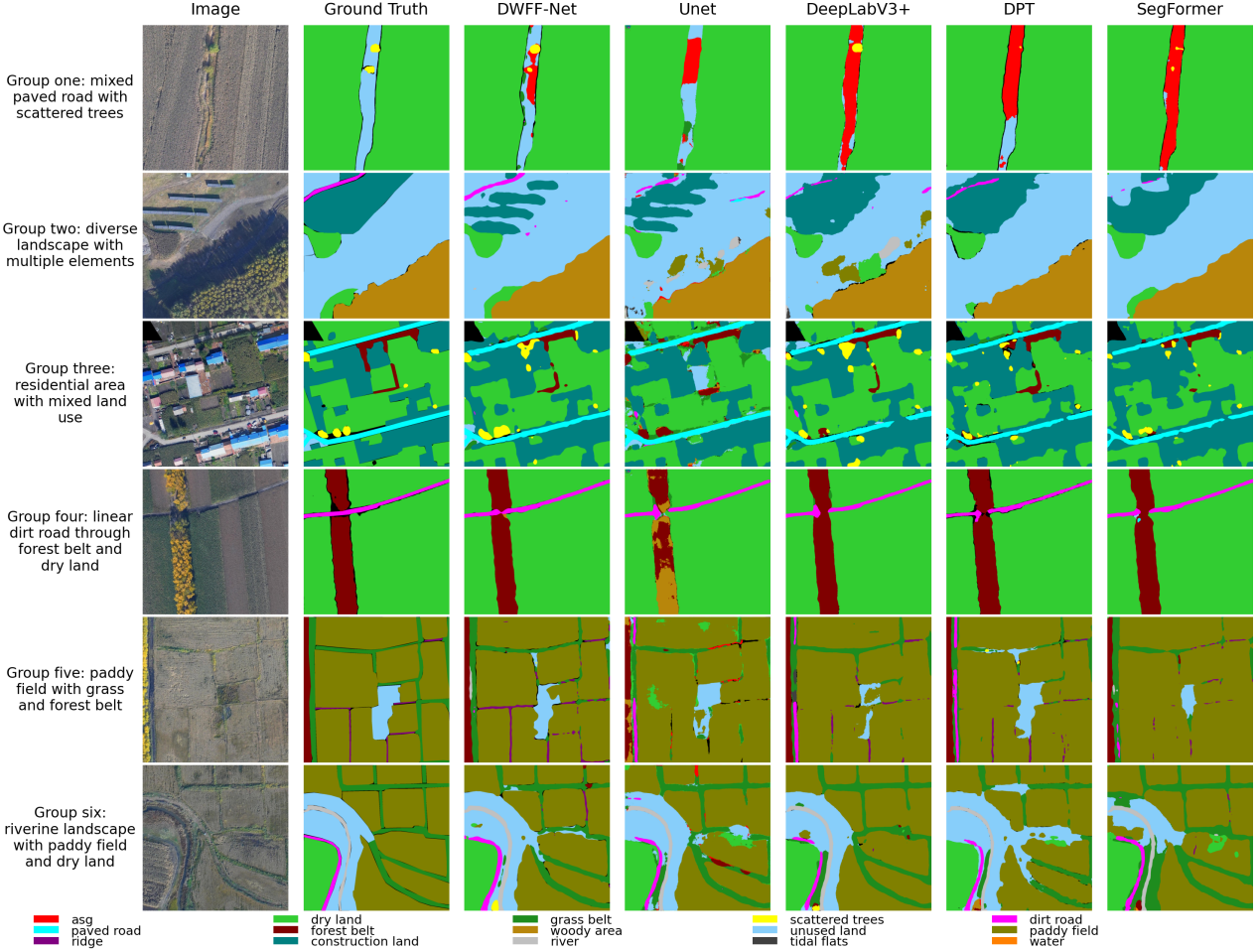}}\hspace{5pt}
\caption{Comparison of extraction effects of U-Net, DeepLa bv3+, DPT, SegFormer and DWFF-Net(ours).} \label{figure15}
\end{figure}

Figure \ref{figure15} shows the visual comparison of the segmentation results obtained by DWFF-Net and several mature segmentation models on test images. In the visual results, it can be observed that the segmentation prediction results of existing methods such as U-Net and DeepLabv3+ have noise and are fragmented severely. The segmentation map generated by our DWFF-Net is exceptionally clear, continuous, and accurate, allowing the topological structure of the field network to be well preserved. For scattered grass belts (Gb) and scattered trees (St) such small target habitats, compared to the models, there are often problems of missed detection, false detection, or merging of adjacent targets. However, DWFF-Net can accurately identify each independent small habitat target and the segmentation boundaries are closer to the actual terrain contour. This further verifies the effectiveness of our dynamic weighted feature fusion strategy, which can better integrate different levels of feature information and effectively improve the segmentation accuracy of small non-cultivated land habitats.

\begin{table}[htbp]
\centering
\caption{IoU values of 15 habitat types in U-Net, DeepLabv3+, DPT, SegFormer and DWFF-Net (ours).}
\label{tab:iou_comparison}
\renewcommand{\arraystretch}{1.2}
\begin{tabular}{lccccc}
\toprule
Class & U-Net & DeepLabv3+ & DPT & SegFormer & DWFF-Net \\
\midrule
Asg   & 0.4243 & \textbf{0.4975} & 0.4275 & 0.4948 & 0.4326 \\
Dl    & 0.9521 & 0.9632 & 0.9640 & 0.9648 & \textbf{0.9682} \\
Gb    & 0.3886 & 0.4568 & 0.3678 & 0.4516 & \textbf{0.4618} \\
St    & 0.0004 & 0.2397 & 0.2184 & 0.2105 & \textbf{0.2707} \\
Dr    & 0.6370 & 0.6236 & 0.6728 & 0.6777 & \textbf{0.6913} \\
Pr    & 0.8276 & 0.8029 & 0.8222 & 0.8298 & \textbf{0.8345} \\
Fb    & 0.6109 & 0.6757 & 0.6484 & 0.6961 & \textbf{0.7008} \\
Wa    & 0.7345 & 0.7853 & 0.7905 & \textbf{0.7952} & 0.7537 \\
Ul    & 0.6279 & 0.6326 & 0.6164 & \textbf{0.6669} & 0.6427 \\
Pf    & 0.9477 & 0.9414 & 0.9663 & 0.9585 & \textbf{0.9741} \\
Ridge & 0.0000 & 0.6353 & 0.6794 & 0.4569 & \textbf{0.6846} \\
Cl    & 0.7960 & 0.8204 & 0.8072 & 0.8036 & \textbf{0.8328} \\
River & 0.5400 & 0.5577 & \textbf{0.7634} & 0.6565 & 0.7626 \\
Tf    & 0.5262 & 0.4529 & 0.4938 & 0.5403 & \textbf{0.5594} \\
Water & 0.7473 & 0.7034 & 0.8976 & 0.8187 & \textbf{0.9090} \\
\bottomrule
\end{tabular}
\end{table}

\section{Conclusion}

This study developed a fine-scale semantic segmentation framework for identifying multi-scale habitats in farmland systems from ultra-high-resolution remote sensing imagery. We divided the agricultural habitat identification dataset constructed in our previous study into training, validation, and test sets in a ratio of 6:1:1 and conducted both ablation and comparative experiments in this dataset. On this basis, a Dynamic-Weighted Feature Fusion Network (DWFF-Net) was proposed by combining a frozen DINOv3 encoder with an adaptive multi-level feature fusion decoder. The proposed strategy dynamically assigns sample-dependent weights to hierarchical features, allowing complementary shallow spatial and texture information and deep semantic representations to be integrated according to the characteristics of individual scenes.

The experimental results demonstrate the effectiveness of the proposed framework. DWFF-Net achieved an mPrecision of 0.8006, mRecall of 0.8131, mF1 of 0.8049, and mIoU of 0.6979, outperforming the non-weighted and static-weighted fusion variants as well as the conventional U-Net, DeepLabv3+, DPT, and SegFormer models in overall segmentation performance. In particular, the dynamic fusion strategy maintained high segmentation accuracy for large and spatially continuous cultivated habitats while providing greater benefits for several small, fragmented, or geometrically complex non-cropped habitats. For example, the IoU for scattered trees reached 0.2707, while the IoUs for tidal flats, rivers, and water bodies reached 0.5594, 0.7626, and 0.9090, respectively. Qualitative comparisons further showed that DWFF-Net produced more continuous predictions, reduced fragmented and missed detections, and better preserved the spatial configuration and boundaries of small habitat patches. The weight-entropy analysis also suggested that the learned fusion weights respond to scene complexity, providing further evidence that adaptive feature weighting can improve the utilization of hierarchical representations for heterogeneous agricultural landscapes.

Despite these improvements, several issues remain to be addressed. The present evaluation was conducted using 0.08m RGB UAV imagery acquired from a single agricultural region in Northeast China and primarily represents autumn habitat conditions; therefore, the transferability of the proposed framework across different seasons, agroecological regions, sensors, and spatial resolutions requires further validation. Moreover, although the identification of small habitats was substantially improved, the relatively low IoU of some highly fragmented classes, particularly scattered trees, indicates that very small targets and complex class boundaries remain challenging. Future research will therefore focus on expanding the dataset across regions and phenological periods, evaluating cross-domain generalization, and further optimizing the dynamic weighting and entropy-regulation mechanisms. Incorporating multi-temporal and multispectral information may also improve the discrimination of spectrally or structurally similar habitats. Overall, the proposed DWFF-Net provides an effective framework for fine-scale farmland habitat mapping and offers a promising technical basis for monitoring non-cropped habitats, assessing agricultural landscape heterogeneity, and supporting biodiversity-oriented farmland management.

\section*{Disclosure statement}

The authors declare that they have no known competing financial interests or personal relationships that could have appeared to influence the work reported in this paper.

\section*{Funding}

This research was supported by the Scientific Research Fund of Liaoning Provincial Education Department of China under [Grant number LJ212510157036].

\section*{Notes on contributors}

Kesong Zheng: Writing -- review \& editing, Data curation, Methodology, Software, Investigation, Formal analysis. Zhi Song: Conceptualization, Formal analysis, Funding acquisition, Investigation, Methodology, Software, Project administration, Writing -- review \& editing. Peizhou Li: Writing -- review \& editing, Writing -- original draft, Data curation,  Software, Visualization, Methodology, Investigation, Formal analysis. Shuyi Yao: Writing -- review \& editing, Methodology, Investigation, Formal analysis. Tong Li: Writing -- review \& editing. Yonglin Shen: Resources. Zhenxing Bian: Writing -- review \& editing, Writing -- original draft, Visualization, Methodology, Investigation, Supervision.

\end{document}